# FP8-RL: A Practical and Stable Low-Precision Stack for LLM Reinforcement Learning

**Zhaopeng Qiu**[1] **Shuang Yu**[1] **Jingqi Zhang**[1] **Shuai Zhang**[1] **Xue Huang**[1]
**Jingyi Yang**[1] **Junjie Lai**[1]

[1]NVIDIA, Beijing, China

alexq@nvidia.com shuangy@nvidia.com larkz@nvidia.com shuazhang@nvidia.com
xueh@nvidia.com sopyang@nvidia.com julienl@nvidia.com

## Abstract

Reinforcement learning (RL) for large language models (LLMs) is increasingly bottlenecked by rollout (generation), where long output sequence lengths make attention and KV cache memory dominate end-to-end step time. FP8 offers an attractive lever for accelerating RL by reducing compute cost and memory traffic during rollout, but applying FP8 in RL introduces unique engineering and algorithmic challenges: policy weights change every step (requiring repeated quantization and weight synchronization into the inference engine) and low-precision rollouts can deviate from the higher-precision policy assumed by the trainer, causing train–inference mismatch and potential instability.

This report presents a practical FP8 rollout stack for LLM RL, implemented in the verl ecosystem with support for common training backends (e.g., FSDP/Megatron-LM) and inference engines (e.g., vLLM/SGLang). We (i) enable FP8 W8A8 linear-layer rollout using blockwise FP8 quantization, (ii) extend FP8 to KV cache to remove long-context memory bottlenecks via per-step QKV scale recalibration, and (iii) mitigate mismatch using importance-sampling-based rollout correction (token-level TIS/MIS variants). Across dense and MoE models, these techniques deliver up to 44% rollout throughput gains while preserving learning behavior comparable to BF16 baselines.

## 1 Introduction

Reinforcement learning (RL) has become a key post-training stage in the development of state-of-the-art large language models (LLMs). However, compared with pretraining and supervised finetuning, RL pipelines impose a distinctive system bottleneck: every optimization step requires fresh rollouts (generation) from the current policy. Recent system analyses show that rollout can dominate end-to-end iteration time—for example, Seer reports the rollout phase consumes approximately 80% of total iteration time in synchronous LLM RL settings [1]. Consequently, accelerating rollout is often the most direct and high-leverage way to improve overall RL throughput and shorten experimentation cycles.

FP8 is an increasingly mature low-precision technology that can accelerate both training and inference on modern GPU architectures, and it is already supported in widely used inference engines (e.g., TensorRT-LLM [2], vLLM [3], and SGLang [4]). Yet applying FP8 in RL is fundamentally more challenging than static inference. First, RL updates policy weights every step, so the rollout engine must repeatedly ingest the newest weights and re-apply quantization at step granularity. This becomes especially challenging in modern RL stacks that decouple training and inference compo-nents and allow independent backend choices, which increases the engineering burden of robust, step-level weight synchronization. Second, low-precision rollouts can introduce training–inference mismatch: updating a high-precision training policy using rollouts generated by a low-precision rollout policy. This effectively introduces an off-policy component, which can substantially destabilize RL and even cause training collapse [5], [6], [7].

This report focuses on building a systematic FP8 solution for LLM RL rollouts, with an emphasis on practicality and reproducibility. We present an FP8 rollout workflow that supports dynamic weight synchronization and quantized loading, and comprehensively evaluate it across model architectures and quantization components in long-context RL training.

Our main contributions are:

1. **Design and implementation of a practical FP8 rollout solution for RL.** We present a production-ready FP8 rollout workflow that handles dynamic weight synchronization and supports multiple training backends (FSDP/Megatron-LM) and inference engines (vLLM/SGLang), addressing the unique engineering challenges of applying FP8 in RL where policy weights change every step.
2. **FP8 quantization for rollout acceleration.** We enable FP8 quantization in rollout through (i) blockwise W8A8 quantization for linear layers to reduce compute cost, and (ii) KV cache quantization via per-step QKV scale recalibration to substantially increase effective cache capacity and concurrency under memory pressure in long-context generation.
3. **Comprehensive validation across model types.** We validate our approach on both dense (Qwen3-8B-Base) and MoE (Qwen3-30B-A3B-Base) models in long-context RL training, demonstrating substantial rollout throughput gains (up to 44% speedup) while maintaining learning behavior comparable to BF16 baselines when paired with standard rollout correction techniques.

## 2 Our Solution: FP8 Rollout for LLM RL

We present an FP8 rollout system for LLM reinforcement learning, implemented within the verl framework with support for common training backends (FSDP/Megatron-LM) and inference engines (vLLM/SGLang). Our approach addresses two core challenges: (i) engineering robust dynamic weight synchronization where policy weights change every step, and (ii) mitigating train-inference mismatch introduced by low-precision rollout.

The system enables FP8 quantization at two levels: **W8A8 quantization for linear layers** to reduce compute cost, and **KV cache quantization** to alleviate memory bottlenecks in long-context generation. Both are paired with importance-sampling-based rollout correction to maintain training stability.

In the following sections, we detail the W8A8 quantization for linear layers (Section 2.1), followed by KV cache quantization (Section 2.3), and conclude with end-to-end FP8 RL exploration (Section 2.4).

### 2.1 W8A8 Quantization for Linear Layers

Our quantization approach operates at three levels:

**Quantization scheme**: We apply blockwise FP8 quantization to linear layers, with weights quantized statically and activations quantized dynamically to balance accuracy and efficiency.

**Weight synchronization**: At each RL step, updated weights from the training backend are quantized and loaded into the inference engine, ensuring the rollout policy reflects the latest training updates.

**Mismatch mitigation**: We employ importance-sampling-based corrections to address the distribution shift between the quantized rollout policy and the full-precision training policy.

#### 2.1.1 Blockwise FP8 Quantization

**Quantization format and granularity.** We adopt the E4M3 FP8 format, which allocates 4 bits to the exponent and 3 bits to the mantissa, providing a dynamic range of approximately [−448, 448] with sufficient precision for neural network computations. To balance quantization error and computational efficiency, we employ blockwise quantization with block size $B = 128 \times 128$ [8]. For a weight matrix $W \in \mathbb{R}^{m \times n}$, we partition it into blocks and compute per-block scaling factors, yielding quantized weights:

$$\widehat{W}_{ij} = \text{round}\left(\frac{W_{ij}}{\text{scale}_{ij}}\right) \in \text{FP8} \tag{1}$$

where $\text{scale}_{ij}$ is derived from the maximum absolute value within each block. This fine-grained approach significantly outperforms per-tensor quantization in preserving model accuracy while remaining hardware-friendly for modern tensor cores.

**Quantization scope.** We apply W8A8 quantization to the following linear layers:

- **Quantized**: attention projections (`q_proj`, `k_proj`, `v_proj`, `o_proj`), MLP layers (`gate_proj`, `up_proj`, `down_proj`), and MoE expert layers (`fc1`, `fc2`).
- **Excluded**: embedding layers, normalization layers, and output projection (`lm_head`). Quantizing `lm_head` may introduce noticeable degradation in generation quality due to its direct impact on vocabulary logits.

**Static vs dynamic quantization.** Weights are quantized statically during the synchronization phase (see below), as they remain constant throughout a rollout batch. Activations, however, vary across inputs and are quantized dynamically during each forward pass to maintain accuracy.

### 2.1.2 Dynamic Weight Synchronization Pipeline

The rollout pipeline follows three phases as illustrated in Figure 1:

- **Initialization phase**: Configure FP8 quantization and apply necessary patches to the inference engine to enable dynamic loading of quantized weights.
- **Weight synchronization phase**: At each RL step, BF16/FP16 weights are retrieved from the training backend (FSDP/Megatron-LM), quantized to FP8 using the blockwise scheme described above, and loaded into the inference engine.
- **Inference phase**: The inference engine executes generation using the pre-loaded FP8 weights. During computation, activation quantization is performed dynamically to maintain accuracy while leveraging FP8 acceleration.

This design supports vLLM and SGLang rollout engines and integrates with common RL training backends (including FSDP and Megatron-LM), with support status tracked in the implementation notes.

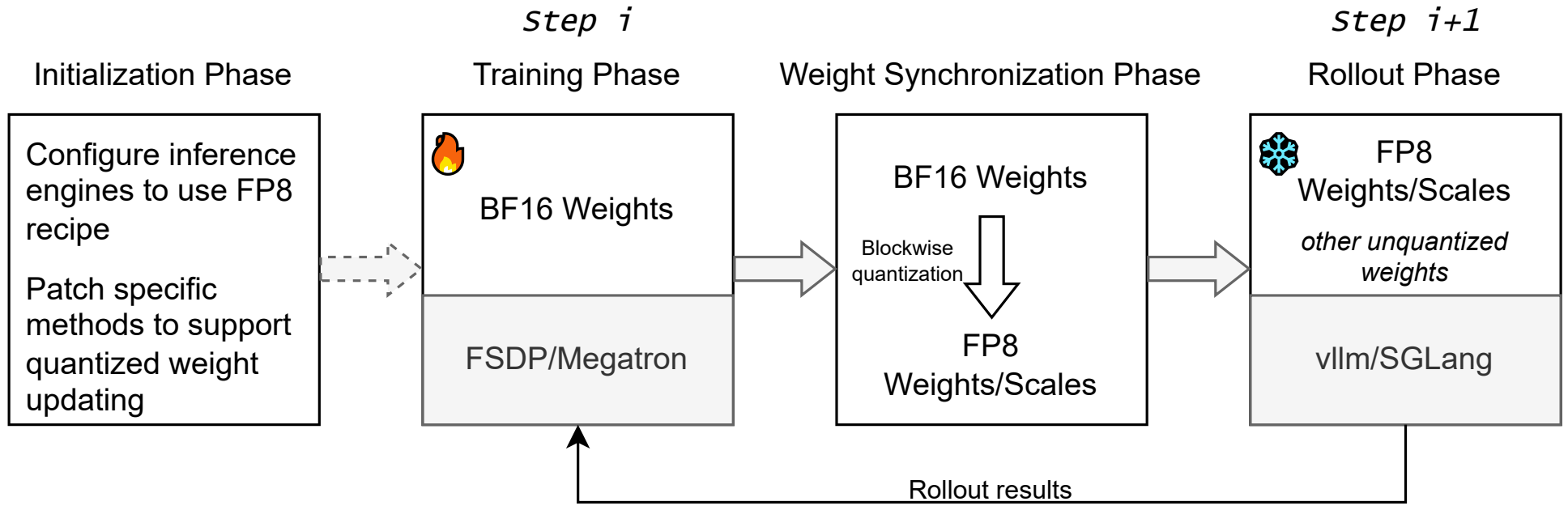

Figure 1: The RL workflow in verl with FP8 W8A8 Linear quantization.

### 2.1.3 Importance Sampling for Rollout Correction

Low-precision rollout introduces a policy mismatch: the training algorithm optimizes $\pi_\theta$ assuming rollouts come from $\pi_\theta$ itself, but samples are actually generated by the quantized policy $\pi_\theta^{\text{FP8}}$. This off-policy component can destabilize training if left unaddressed.

**Importance weighting.** We correct this mismatch using importance sampling, which reweights each action $a$ by the likelihood ratio:

$$w(a|s) = \frac{\pi_\theta(a|s)}{\pi_\theta^{\text{FP8}}(a|s)} \tag{2}$$

In practice, we compute these ratios at token level during training, using the BF16 policy $\pi_\theta$ to evaluate the log-probabilities of tokens sampled by the FP8 rollout policy $\pi_\theta^{\text{FP8}}$.

**Truncated importance sampling (TIS).** Naive importance weights can exhibit high variance, particularly when $\pi_\theta$ and $\pi_\theta^{\text{FP8}}$ diverge due to quantization error accumulated over long sequences. We adopt token-level truncated importance sampling (TIS), which clips the importance weights to a bounded range:

$$w_{\text{TIS}}(a|s) = \text{clip}(w(a|s), C) \tag{3}$$

where $C$ is the clipping threshold (we use $C = 2$ in all experiments). This bounded correction stabilizes training while still correcting for the primary distribution shift.

### 2.1.4 Implementation

Our FP8 W8A8 linear rollout is implemented within the verl framework [9] and can be enabled by setting `actor_rollout_ref.rollout.quantization=fp8` in the training configuration. The system automatically handles weight conversion and loading into the inference backend. For optimal performance, we recommend using CUDA 12.9+ and enabling the DeepGEMM[1] library for accelerated FP8 GEMM operations (enabled by default in vLLM 0.11+ and SGLang 0.55+). It is strongly recommended to pair FP8 rollout with importance-sampling-based rollout correction (e.g., token-level TIS) to maintain training stability. Detailed configuration examples and usage instructions are provided in Section A.

## 2.2 Experiments

In this section, we detail our experimental setup and present results for both dense and MoE models, demonstrating the efficacy of FP8 W8A8 linear rollout in end-to-end RL training scenarios.

### 2.2.1 Settings

**Training Setup** We evaluate our approach using the DAPO [10] algorithm with online validation on the AIME24 benchmark. All experiments are conducted on H100-class GPUs, using 8×H100 for the 8B dense model and 2×8×H100 for the 30B MoE model. We use vLLM as the rollout engine and FSDP as the training backend.

**Hyperparameters** We use a prompt batch size of 32 with n=16 responses per prompt, resulting in a rollout batch size of 32×3×16. Both training batch size and PPO mini-batch size are set to 32 to ensure the policy is updated using rollout outputs only once per iteration, which helps isolate the impact of quantization by removing additional off-policy noise. The maximum response length is set to 20K tokens to stress-test the system under long-context generation scenarios. To mitigate train-inference mismatch introduced by FP8 quantization, we apply token-level truncated importance sampling (TIS) with clipping threshold C=2 for all FP8 configurations. While RL training hyperparameters are highly case-dependent, these settings represent a verified working recipe for the models tested. Complete configuration details, including software versions and training scripts, are provided in Section A.

[1] https://github.com/deepseek-ai/DeepGEMM

**Metrics** We evaluate training effectiveness and system performance using the following metrics: (i) *validation accuracy* on the AIME24 test set, measuring the percentage of correctly solved problems as the primary indicator of policy learning quality; (ii) *reward*, the average reward per response computed by the reward model during training, tracking optimization progress; (iii) **response length**, the average token count per generated response, which typically increases as the policy learns more complex reasoning chains in long-context RL; (iv) *mismatch KL*, the KL divergence $D_{\mathrm{KL}}\left(\pi_\theta^{\mathrm{FP8}} \parallel \pi_\theta\right)$ between the rollout policy $\pi_\theta^{\mathrm{FP8}}$ (inference) and training policy $\pi_\theta$, quantifying the distribution shift introduced by quantization—lower values indicate better alignment, while elevated values can destabilize training if left uncorrected; and (v) *rollout performance*, measured via time-per-token (ms/token) during generation, where lower values indicate faster rollout and reduced end-to-end training iteration time.

### 2.2.2 Dense Model: Qwen3-8B-Base

We conduct Qwen3-8B-Base model based experiments under three configurations: (i) BF16 baseline without rollout correction (orange), (ii) FP8 W8A8 with token-level TIS (blue), and (iii) FP8 W8A8 without TIS (green) for ablation analysis. Figure 2 presents the training dynamics across all configurations.

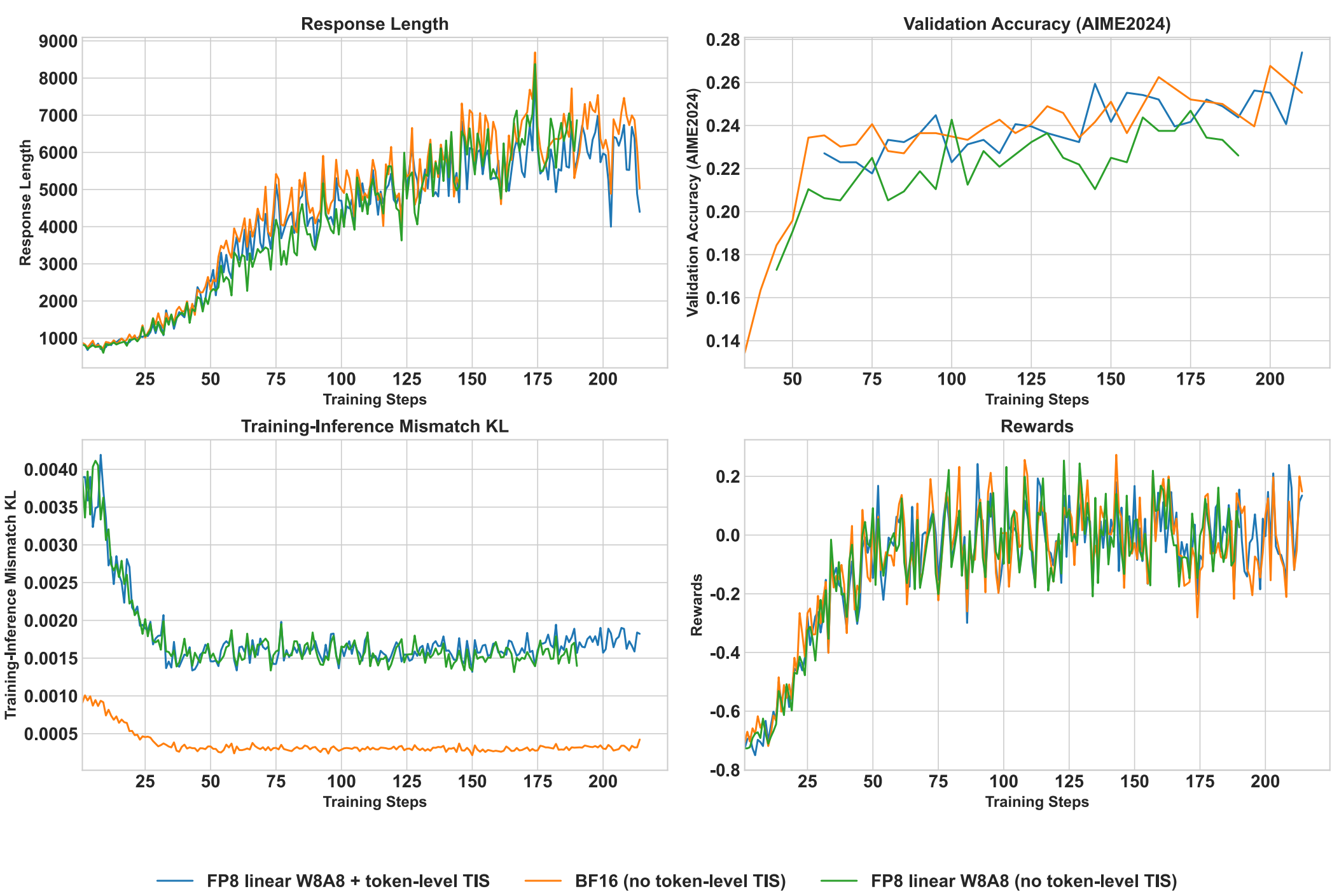

Figure 2: Training curves for Qwen3-8B-Base. Orange: BF16 baseline without TIS. Blue: FP8 W8A8 with token-level TIS. Green: FP8 W8A8 without TIS.

**Training effectiveness.** Our solution, FP8 W8A8 with token-level TIS (blue), closely tracks the BF16 baseline (orange) across all key training metrics. Validation accuracy curves align well throughout training, with both configurations achieving comparable final accuracy on the AIME24 benchmark. Reward curves overlap significantly, confirming that policy optimization proceeds effectively under FP8 rollout. Response length grows consistently in both configurations, indicating that the policy successfully learns to generate increasingly complex reasoning chains regardless of rollout precision. The mismatch KL, which quantifies the distribution shift between the rollout policy $\pi_\theta^{\mathrm{FP8}}$ and the training policy $\pi_\theta$, is naturally higher for the FP8 run due to quantization-induced precision loss. However, with TIS enabled, this divergence remains stable and within an acceptable range, ensuring training stability while maintaining accuracy comparable to the BF16 baseline.

**Importance of rollout correction.** The ablation configuration—FP8 W8A8 without TIS (green) —exhibits noticeable accuracy degradation compared to both the BF16 baseline and the FP8+TIS configuration. This performance drop demonstrates the critical role of importance sampling in mitigating the distribution mismatch introduced by quantization.

**Rollout performance.** We evaluate the system-level performance impact of FP8 W8A8 linear rollout using CUDA 12.9 to maximize DeepGEMM kernel efficiency. Figure 3 shows time-per-token measurements across varying response lengths. FP8 rollout consistently outperforms BF16 across all sequence lengths, yielding an overall speedup of approximately 10–20% for the 8B dense model. Performance gains are more pronounced at longer sequence lengths, where memory bandwidth constraints become the dominant bottleneck and FP8′s reduced memory traffic provides greater benefit.

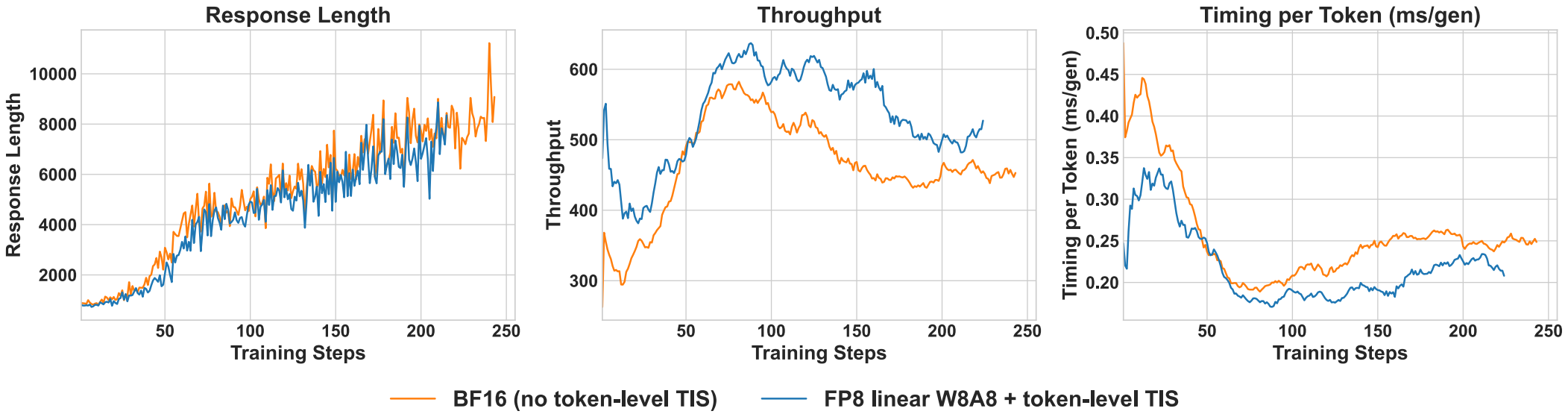

Figure 3: Rollout performance on Qwen3-8B-Base. Time-per-token (ms/token) during generation and throughput for varying response lengths. Blue: FP8 W8A8 linear rollout. Orange: BF16 baseline.

### 2.2.3 MoE Model: Qwen3-30B-A3B-Base

We evaluate FP8 W8A8 linear rollout on the Qwen3-30B-A3B-Base MoE model, comparing two configurations: (i) BF16 with token-level TIS (orange) and (ii) FP8 W8A8 with token-level TIS (blue). Unlike the dense model experiments, we apply rollout correction to both configurations because MoE models exhibit more pronounced train-inference mismatch even at full precision, making correction beneficial for optimal convergence. Figure 4 presents the training dynamics for both configurations.

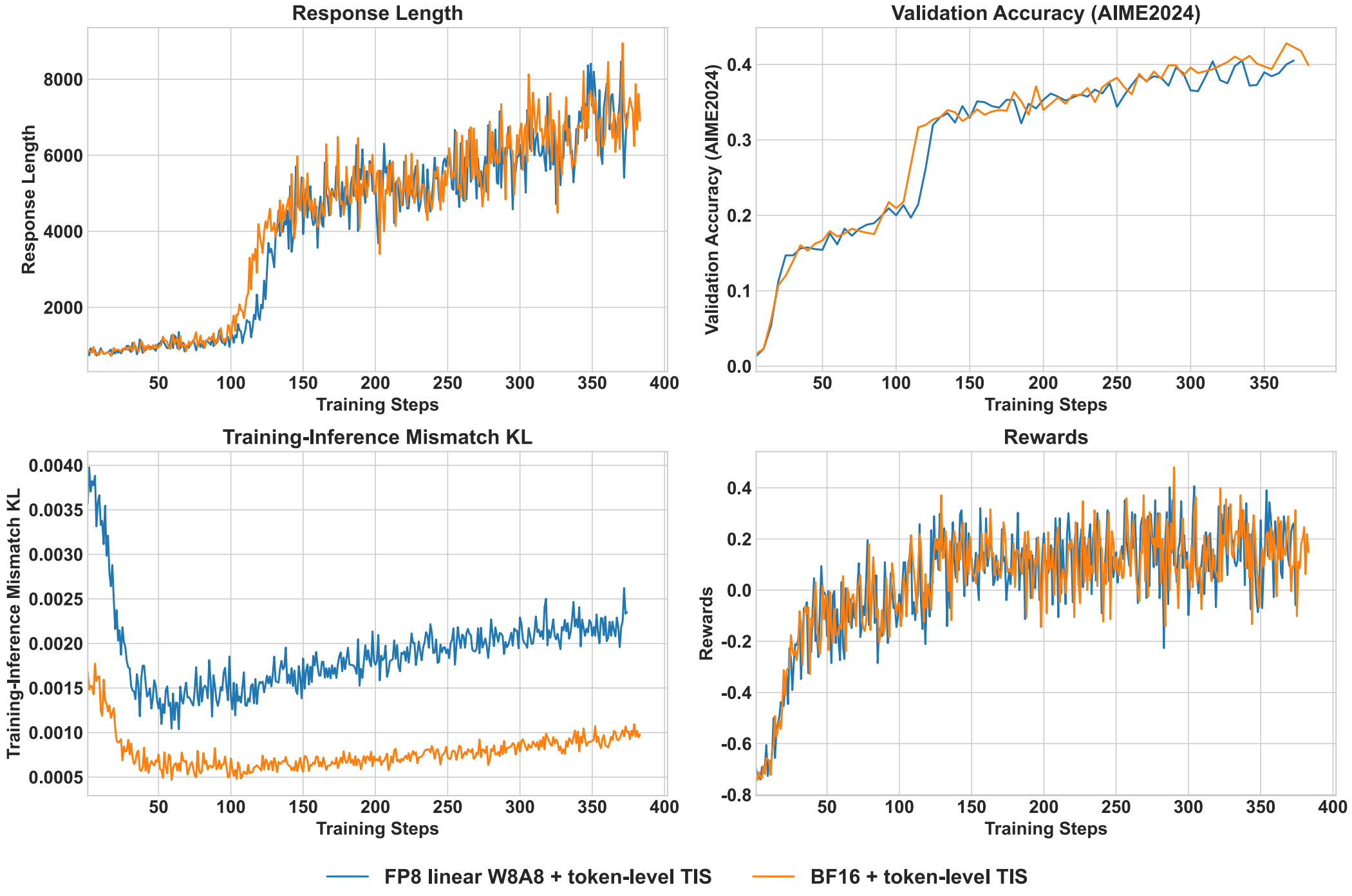

Figure 4: Training curves for Qwen3-30B-A3B-Base MoE. Orange: BF16 with token-level TIS. Blue: FP8 W8A8 with token-level TIS.

**Training effectiveness.** With token-level TIS enabled, the FP8 rollout (blue) aligns remarkably well with the BF16 baseline (orange) across all key training metrics. Validation accuracy curves track closely throughout training, demonstrating that FP8 quantization preserves learning quality for MoE models. Reward and response length curves also overlap significantly, confirming that policy optimization proceeds effectively under FP8 rollout despite the additional complexity introduced by mixture-of-experts routing.

**MoE-specific observations.** Unlike the dense model where mismatch KL remains stable, MoE models inherently exhibit a trend of increasing mismatch KL during training in both the BF16 and FP8 runs. This phenomenon stems from the dynamic nature of MoE architectures, where slight differences in precision or implementation between the inference engine and training backend can lead to inconsistent expert selection across the two systems. As training progresses and the policy evolves, these routing inconsistencies accumulate, causing the mismatch KL to gradually increase. While token-level TIS successfully kept this divergence manageable in our experiments—as evidenced by the aligned validation accuracy—scenarios with more severe instability may require advanced correction techniques such as Masked Importance Sampling (MIS) or Rollout Router Replay (RRR), which enforces consistent expert selection between rollout and training.

**Rollout performance.** FP8 W8A8 linear rollout delivers substantially larger performance gains for the MoE model compared to the dense model. As shown in Figure 5, FP8 rollout (blue) achieves significantly lower time-per-token compared to BF16 (orange), yielding an overall speedup of approximately 30–50% across varying response lengths. This represents a 2–3× larger improvement compared to the 10–20% speedup observed for the 8B dense model.

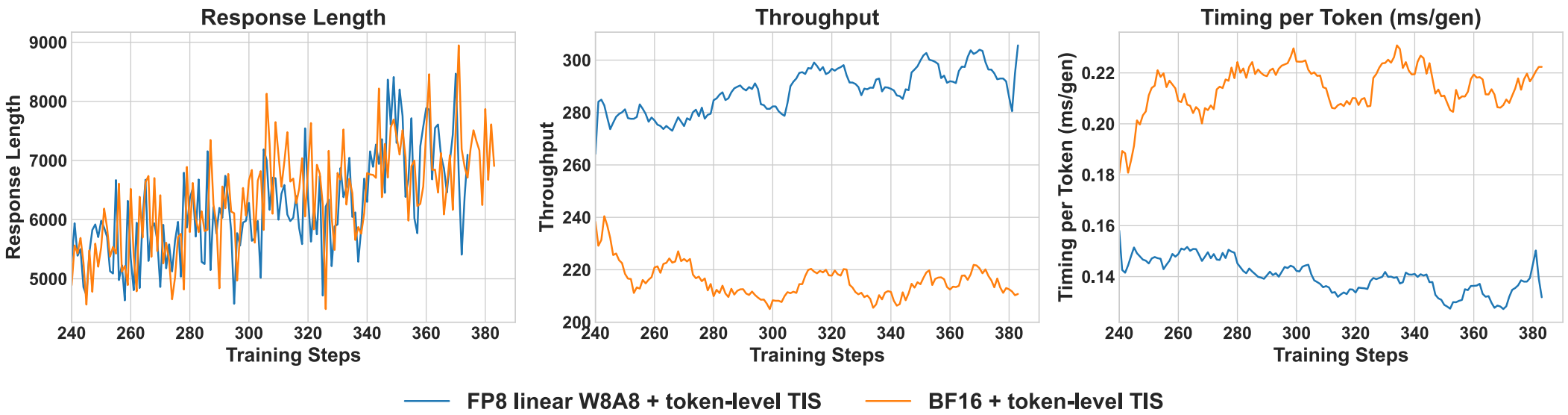

Figure 5: Rollout performance on Qwen3-30B-A3B-Base MoE. Time-per-token (ms/token) during generation for varying response lengths. Blue: FP8 W8A8 linear rollout. Orange: BF16 baseline.

**Performance analysis.** The substantially larger speedup for the 30B MoE model can be attributed to three complementary factors. First, larger models exhibit higher arithmetic intensity, making FP8's compute acceleration more impactful—the 30B model's matrix multiplications are significantly larger than those in the 8B model, allowing FP8 tensor cores to deliver greater absolute time savings. Second, reduced memory traffic becomes increasingly beneficial at scale: loading and transferring the massive 30B parameter set consumes substantial bandwidth in BF16, and FP8's 2× memory footprint reduction directly translates to faster weight loading and reduced memory pressure. Third, quantizing the 30B weights frees significant GPU memory, which expands the available KV cache capacity. This expanded headroom allows the system to host more concurrent requests and tokens, reducing preemption frequency and increasing overall throughput. These factors compound in long-context generation scenarios where both compute intensity and memory pressure are high, explaining why the 30B MoE model benefits disproportionately from FP8 W8A8 linear rollout compared to smaller dense models.

### 2.2.4 Ablation: Router Precision for MoE FP8 Rollout

MoE routing decisions are particularly sensitive to numerical precision, as small perturbations in router logits can alter expert selection and cascade into divergent model behavior. To isolate the effect of router precision on rollout-side mismatch, we fix training in BF16 and vary the router precision during FP8 rollout among three settings: FP8 (router quantized along with other layers), BF16 (router excluded from quantization), and FP32 (router explicitly upcast to FP32 on both the vLLM and Megatron-Core sides). Figure 6 presents mismatch KL and validation accuracy for the first 200 training steps.

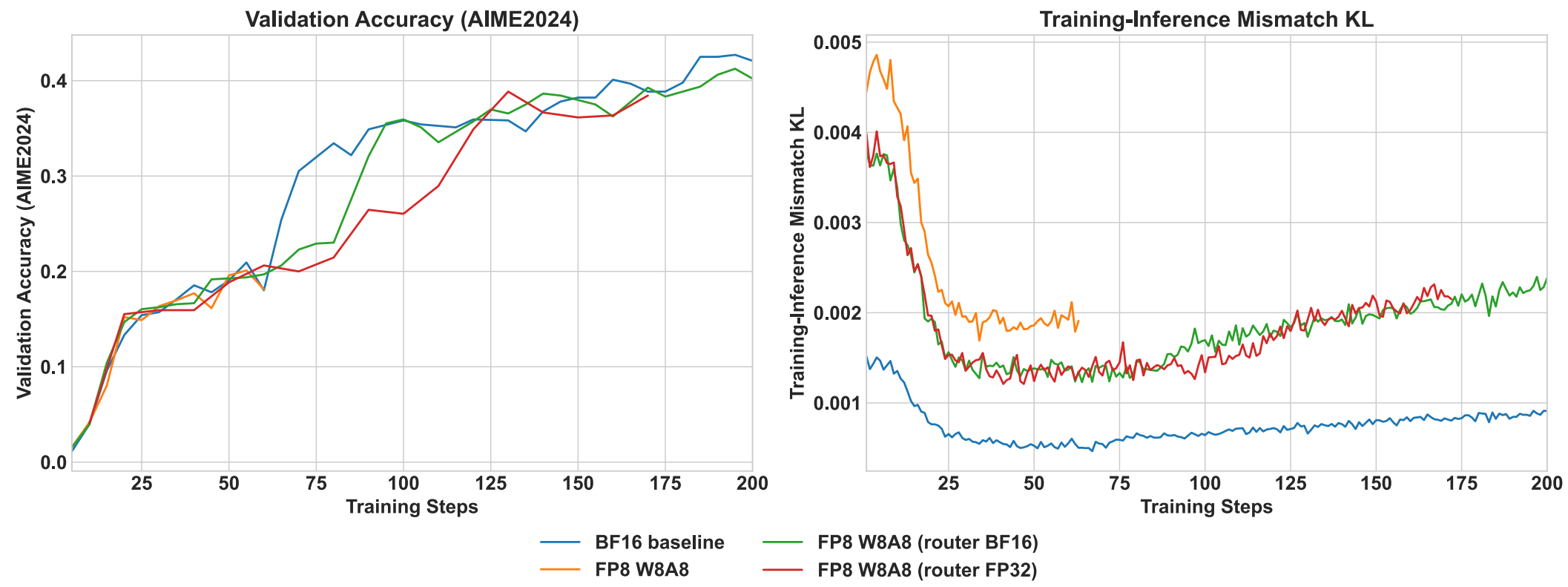

Figure 6: Ablation on router precision during FP8 rollout for Qwen3-30B-A3B-Base (BF16 training). Blue: BF16 baseline. Orange: FP8 rollout with router in FP8. Green: FP8 rollout with router in BF16. Red: FP8 rollout with router in FP32 (both vLLM and Megatron-Core sides).

The FP8 router (orange) exhibits the highest mismatch KL, particularly in the early training phase (~0.004), reflecting greater routing inconsistency between training and rollout when the router operates at the lowest precision. The BF16 router (green) and FP32 router (red) show similar and lower mismatch KL, both closer to the BF16 baseline (blue). This suggests that excluding the router from FP8 quantization is sufficient to maintain routing consistency—upgrading further from BF16 to FP32 provides marginal additional benefit.

### 2.3 Beyond Linear Layers: FP8 KV Cache Quantization

Because long-context RL rollouts are often memory-bound, KV cache quantization is a high-impact extension beyond linear layer only quantization, targeting the dominant attention/memory bottleneck during generation.

#### 2.3.1 Implementation Strategy: Dynamic QKV Scale Recalibration

To enable FP8 KV cache in an RL setting, the key challenge is handling dynamic weights. Unlike static inference, the policy changes every step, meaning QKV scales must be recalibrated frequently. We explored two design paradigms to address this:

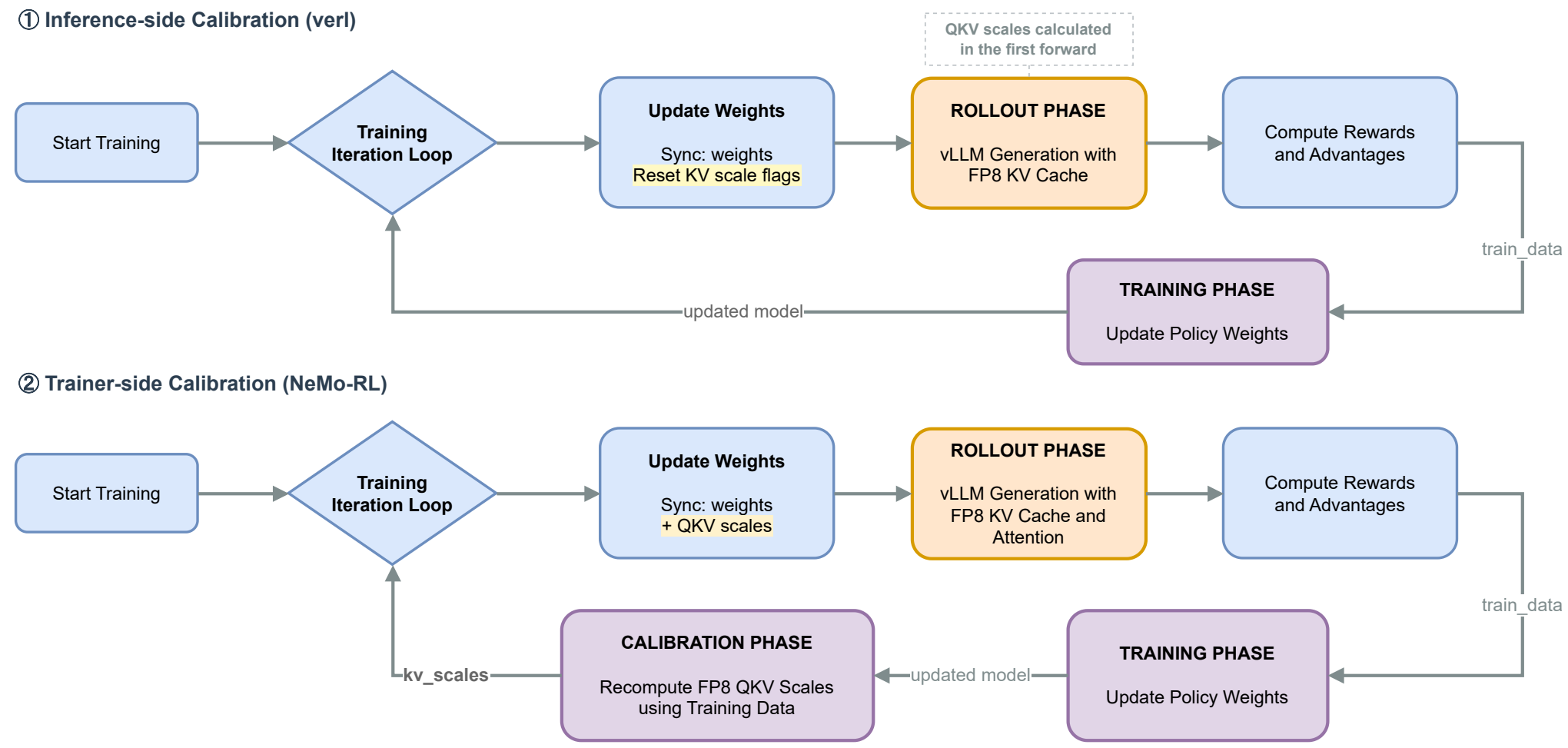

Figure 7: Two implementation strategies for FP8 KV cache.

- **Inference-Side calibration**: Modern inference engines (like vLLM) natively support dynamic QKV scale calculation, typically performing this calibration during the first forward pass after initialization. As shown in Figure 7, this approach leverages that built-in capability but adapts it for RL: we trigger a forced recalibration (e.g., by resetting the internal `calculate_kv_scales` flags of all the attention layers) before the rollout phase of each RL step. This ensures that the KV cache quantization accurately reflects the newly updated policy weights from the previous training step without external intervention.

- **Trainer-Side calibration**: As depicted by Figure 7, at the end of each training step, the training backend recalibrates QKV scales using the updated policy weights and a subset of training data (prompts and generated responses). These new scales are then synchronized to the inference engine for the subsequent rollout. This method introduces tighter inference backend coupling and minor calibration overhead but offers fine-grained control over the calibration data and process.

We implemented Inference-Side calibration in verl and Trainer-Side calibration in NeMo-RL [11]. In the following experiments, we present results from the verl implementation. Usage instructions for both frameworks and additional experimental validation of the NeMo-RL implementation are provided in Section B.

#### 2.3.2 Experiments

We evaluate KV cache quantization on Qwen3-8B-Base using the verl implementation with Inference-Side calibration. The experimental setup follows the same DAPO recipe and H100 hardware configuration as the W8A8 linear rollout experiments. We compare four configurations:

- **BF16 baseline**: Full BF16 precision throughout.
- **Linear W8A8**: W8A8 quantization for linear layers only, as described in Section 2.1.
- **KV cache FP8 only**: FP8 quantization applied exclusively to KV cache storage, while linear layers and attention computations remain in BF16. This isolates the impact of KV cache quantization.
- **Full FP8**: FP8 quantization across all three components—linear layers (W8A8), KV cache storage, and attention computations—representing the most aggressive quantization setting.

All FP8 configurations use token-level TIS (C=2) for rollout correction and target a maximum response length of 20K tokens. Figure 8 presents the training dynamics.

**Training effectiveness.** KV cache FP8 only aligns closely with the BF16 baseline on validation accuracy, indicating that quantizing the KV cache alone has minimal impact on model convergence. Reward curves and response length growth also track the baseline, confirming that policy optimization proceeds effectively despite KV cache quantization.

The KV cache FP8 only run exhibits a slightly higher mismatch KL compared to the Linear W8A8 run. We hypothesize this stems from error accumulation in long-context generation: unlike weight quantization which is static per layer, KV cache quantization affects every token's attention computation dynamically over the entire long sequence. However, despite this slight increase in KL divergence, validation accuracy and reward remain stable and aligned with the baseline, demonstrating that token-level TIS successfully mitigates this mismatch.

Full FP8, which combines linear, KV cache, and attention quantization, shows the largest mismatch KL as expected from compounded quantization errors. Nevertheless, with token-level TIS enabled, it still follows the general accuracy trend, confirming that even aggressive full-stack quantization can be viable for RL training when paired with robust mismatch mitigation.

**Rollout performance.** Figure 9 shows rollout performance across different quantization configurations. Linear W8A8 achieves approximately 20% speedup over the BF16 baseline, while KV cache FP8 only delivers a substantially larger 38% speedup. Combining both linear and KV cache quantization yields a cumulative 44% speedup, demonstrating the complementary benefits of quantizing both components.

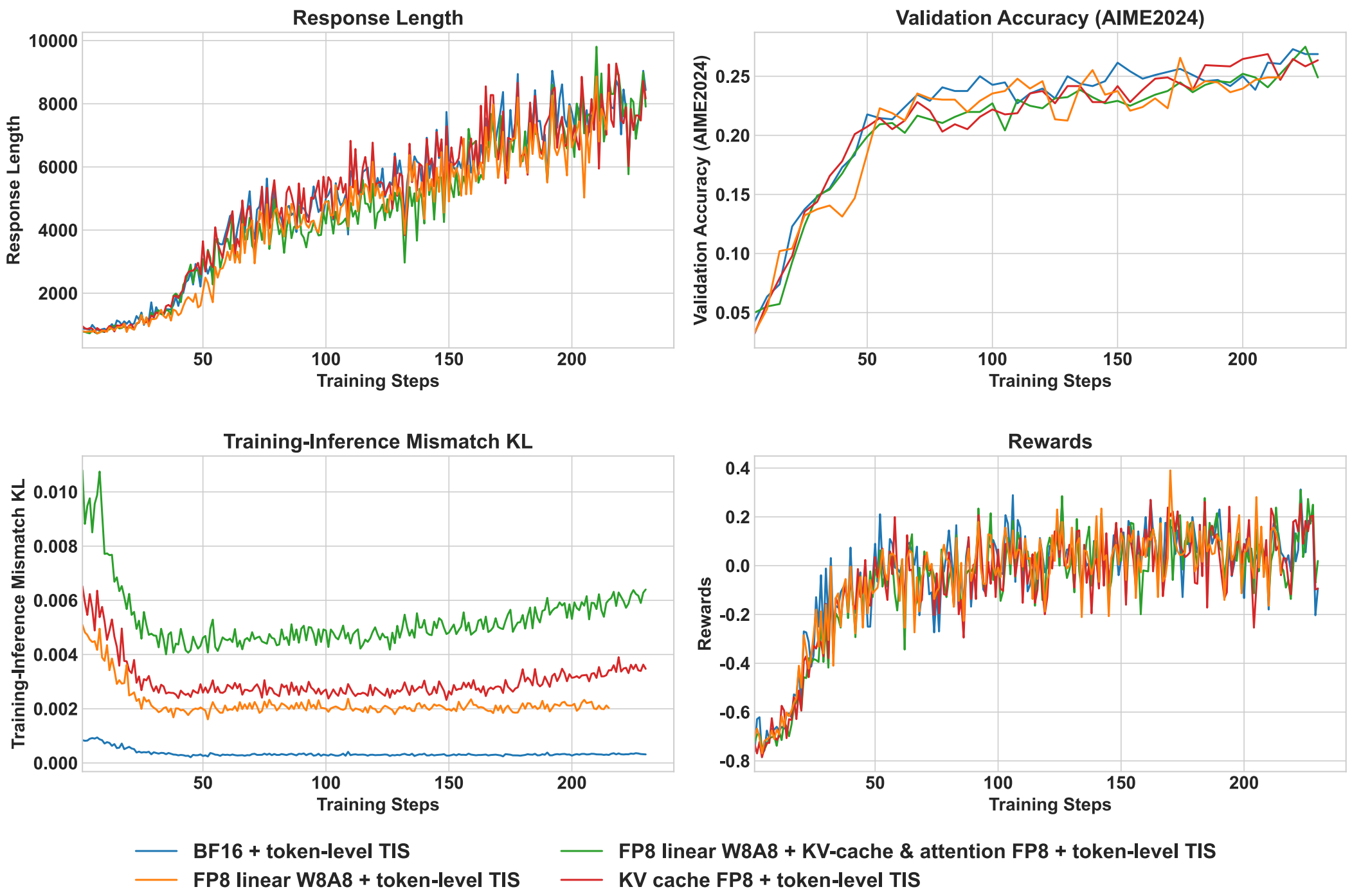

Figure 8: Training curves for Qwen3-8B-Base with KV cache FP8. Blue: BF16 baseline. Orange: Linear W8A8 + TIS. Red: KV cache FP8 only + TIS. Green: Full FP8 (Linear + KV + Attention) + TIS.

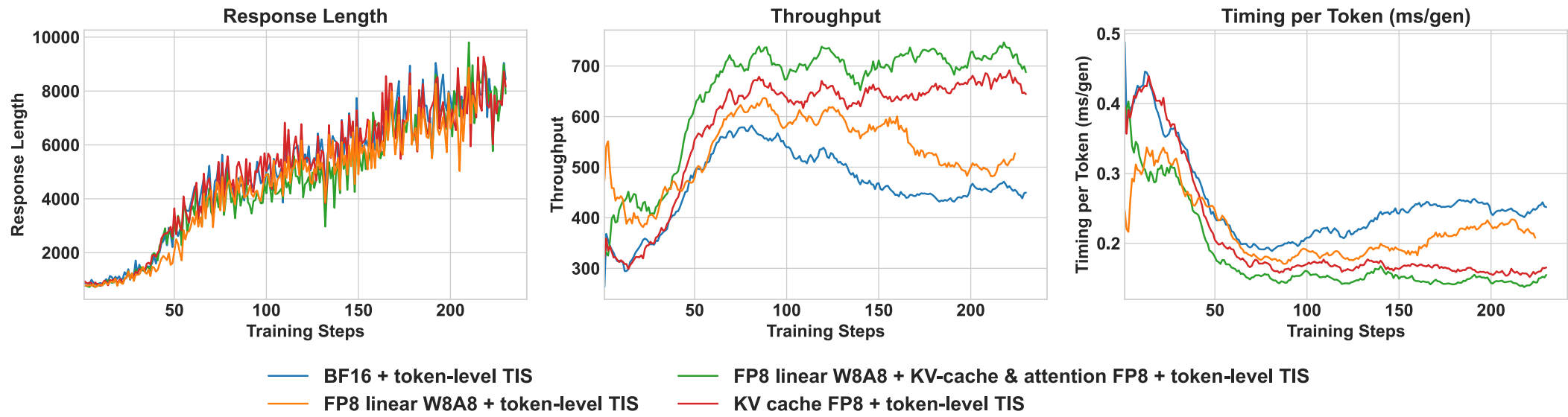

Figure 9: Rollout speedup on Qwen3-8B-Base. Blue: BF16 baseline. Orange: Linear W8A8 + TIS. Red: KV cache FP8 only + TIS. Green: Full FP8 (Linear + KV + Attention) + TIS.

**Performance analysis.** KV cache quantization provides disproportionately large gains ( 38%) compared to linear quantization alone ( 20%) in this experiment. This outcome is highly dependent on model size and use case: for small dense models (like 8B) generating very long responses (maximum 20K tokens), the workload becomes heavily memory-bound and bottlenecked on KV cache capacity rather than compute. In our experiments with the BF16 baseline and linear W8A8 configurations, vLLM monitoring revealed frequent request preemptions caused by insufficient KV cache space, which wasted computation and throttled throughput. By enabling KV cache FP8, we effectively double the KV cache capacity, increasing maximum concurrency and reducing preemption frequency. This expanded headroom significantly mitigates memory pressure, resulting in a stable, high-throughput generation stream that maximizes GPU utilization.

### 2.4 End-to-End FP8 RL

The preceding sections evaluated FP8 rollout in isolation, maintaining BF16 precision during training. While this configuration already delivers substantial system-level gains, applying FP8 end-to-end—across both rollout and training—offers three additional benefits. First, FP8 training has been validated at scale in pretraining and supervised finetuning, where it matches BF16 convergence when using appropriate formats and scaling rules. Second, end-to-end FP8 can reduce precision-induced mismatch between training and inference: rollout-only FP8 keeps training in BF16/FP16 while generating trajectories in FP8, amplifying distribution drift due to quantization/dequantization error and kernel-level differences. By aligning the precision of both paths, end-to-end FP8 makes the distribution optimized during learning closer to the distribution produced during rollout. Third, FP8 training can accelerate the RL loop by speeding up the learner-side forward/backward passes for policy and value networks, improving time-to-train and reducing end-to-end cost beyond rollout-only acceleration.

We validate end-to-end FP8 on the Qwen3-30B-A3B-Base MoE model and conduct ablation studies on key design choices—router precision, FP8 recipe, and scaling factor format. Additional validation on the Qwen3-8B-Base dense model is provided in Section C.

### 2.4.1 Settings

**Experimental setup.** We evaluate end-to-end FP8 on the Qwen3-30B-A3B-Base MoE model using the DAPO algorithm with AIME24 online validation. Experiments are conducted on 4×8×H100 GPUs using verl with vLLM 0.12.0 as the rollout engine and Megatron-Core 0.15 as the training backend. We use a prompt batch size of 128 with $n = 16$ responses per prompt (rollout batch size $128 \times 3 \times 16$), and maximum response length of 20K tokens. The FP8 training uses the hybrid recipe (E4M3 forward, E5M2 backward), and the MoE router is kept in BF16 on both the vLLM and Megatron-Core sides to preserve routing stability. Token-level TIS with C=2 is applied to all configurations for rollout correction.

**Configurations.** We compare three precision settings:

- **BF16 training + BF16 rollout**: Full BF16 baseline
- **BF16 training + FP8 rollout**: FP8 rollout-only (the configuration from Section 2.1)
- **FP8 training + FP8 rollout**: End-to-end FP8 quantization

### 2.4.2 Results

Figure 10 presents the training dynamics across all three configurations, including validation accuracy, response length, reward, and mismatch KL.

**Accuracy alignment with BF16.** FP8 end-to-end (orange) closely tracks the BF16 baseline (purple) throughout training, with both configurations reaching comparable final accuracy (~0.55–0.6 on AIME24). Reward curves and response length growth also align well, confirming that end-to-end FP8 preserves the learning dynamics of the BF16 run on MoE models.

**Mismatch analysis.** The mismatch KL reveals a clear ordering across configurations: FP8 rollout-only (green) > FP8 end-to-end (orange) > BF16 (purple). FP8 rollout-only exhibits progressively increasing mismatch KL, culminating in a sharp spike around step 700 that drives the KL above 5. This spike coincides with a noticeable drop in both accuracy and response length, effectively crashing the training run. FP8 end-to-end also shows a gradually increasing mismatch KL trend over training, but the values remain within a reasonable range and the long-term trend aligns with that of the BF16 baseline. This confirms that aligning precision between training and rollout substantially reduces distribution drift, though it does not fully eliminate mismatch—residual sources such as kernel-level numerical differences between the training backend and inference engine persist.

**MoE-specific instability under mixed precision.** The crash of the FP8 rollout-only configuration echoes the MoE-specific mismatch observations from Section 2.1: MoE models are inherently more vulnerable to train–inference divergence due to discrete expert routing, and mixed-precision operation amplifies this effect. When the training backend operates in BF16 while rollout uses FP8, routing inconsistencies accumulate over training steps and eventually trigger catastrophic divergence. We note that in these experiments, token-level TIS is the only mismatch mitigation applied—no additional techniques such as Rollout Router Replay (R3) or Masked Importance Sampling (MIS) are used. The FP8 rollout-only crash suggests that for MoE models, TIS alone may be insufficient to control mismatch under mixed precision, and stronger correction mechanisms may be necessary in that setting.

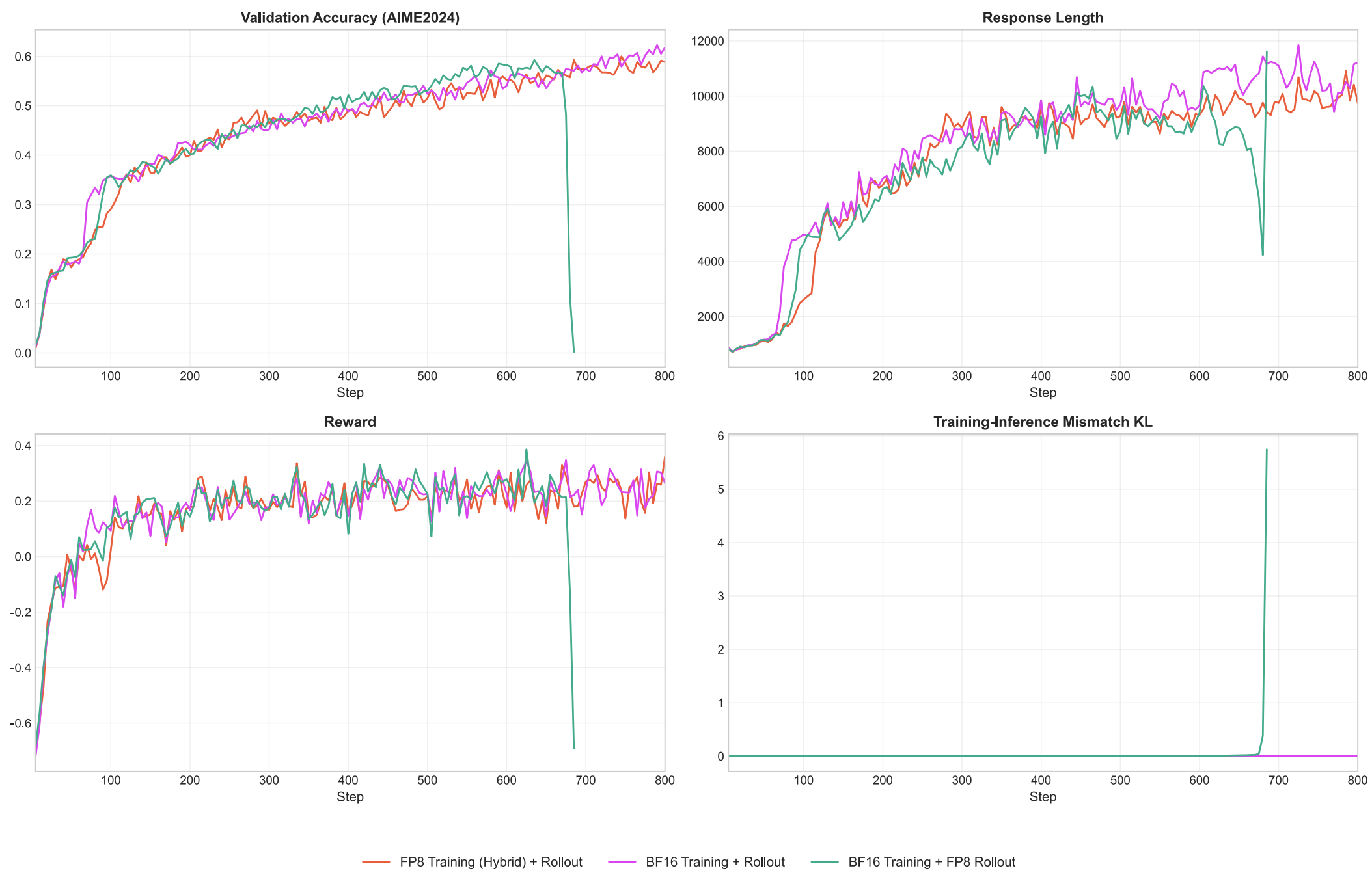

Figure 10: Training curves for end-to-end FP8 RL on Qwen3-30B-A3B-Base MoE. Purple: BF16 training + BF16 rollout. Orange: FP8 training + FP8 rollout. Green: BF16 training + FP8 rollout.

**Training-side performance.** In our current experiments, we did not observe significant training-side speedup from enabling FP8 training on the 30B MoE model. While FP8 GEMM operations are theoretically faster, the RL training loop involves substantial non-GEMM overhead (e.g., data loading, reward computation, weight synchronization) that dilutes the per-operator speedup. The primary benefit of end-to-end FP8 in our setup is the reduction in train–inference mismatch rather than training acceleration. We expect training-side gains to become more pronounced as the training workload scales further (e.g., larger batch sizes or longer sequences that increase the compute-bound fraction).

### 2.4.3 Ablation Studies on Qwen3-30B-A3B-Base

We conduct ablation studies on key design choices for end-to-end FP8 training on the Qwen3-30B-A3B-Base MoE model, focusing on the FP8 recipe and scaling factor format. All ablations share the same base configuration as Section 2.4.1 unless otherwise specified.

**FP8 Training Recipe: Hybrid vs E4M3.**

The FP8 hybrid training recipe uses E4M3 for forward passes and E5M2 for backward passes, balancing precision in activations with dynamic range in gradients. An alternative approach, following DeepSeek-V3 [8], uses E4M3 for both forward and backward passes. We compare these two recipes in end-to-end FP8 RL. Figure 11 presents training dynamics across six metrics.

The hybrid recipe (orange) closely tracks the BF16 baseline (purple) across all metrics—validation accuracy, reward, response length, entropy, and mismatch KL—throughout the full 600-step training run. In contrast, the pure E4M3 recipe (green) trains stably for the first ~500 steps but then collapses: the gradient norm spikes sharply (from ~0.3 to ~1.75), accompanied by a surge in entropy and mismatch KL ( 0.25). This cascading instability leads to a rapid drop in accuracy and response length.

This failure is consistent with the known limitation of E4M3 for gradient representation: E4M3 provides only 4 exponent bits, giving a dynamic range of $[-448, 448]$, whereas E5M2 provides 5 exponent bits with a range of $[-57344, 57344]$. During RL training, gradient magnitudes can vary widely—particularly in later training stages where the policy has learned long reasoning chains and gradient signals become more heterogeneous. The limited dynamic range of E4M3 leads to gradient underflow or loss of precision in the backward pass, eventually triggering the observed instability.

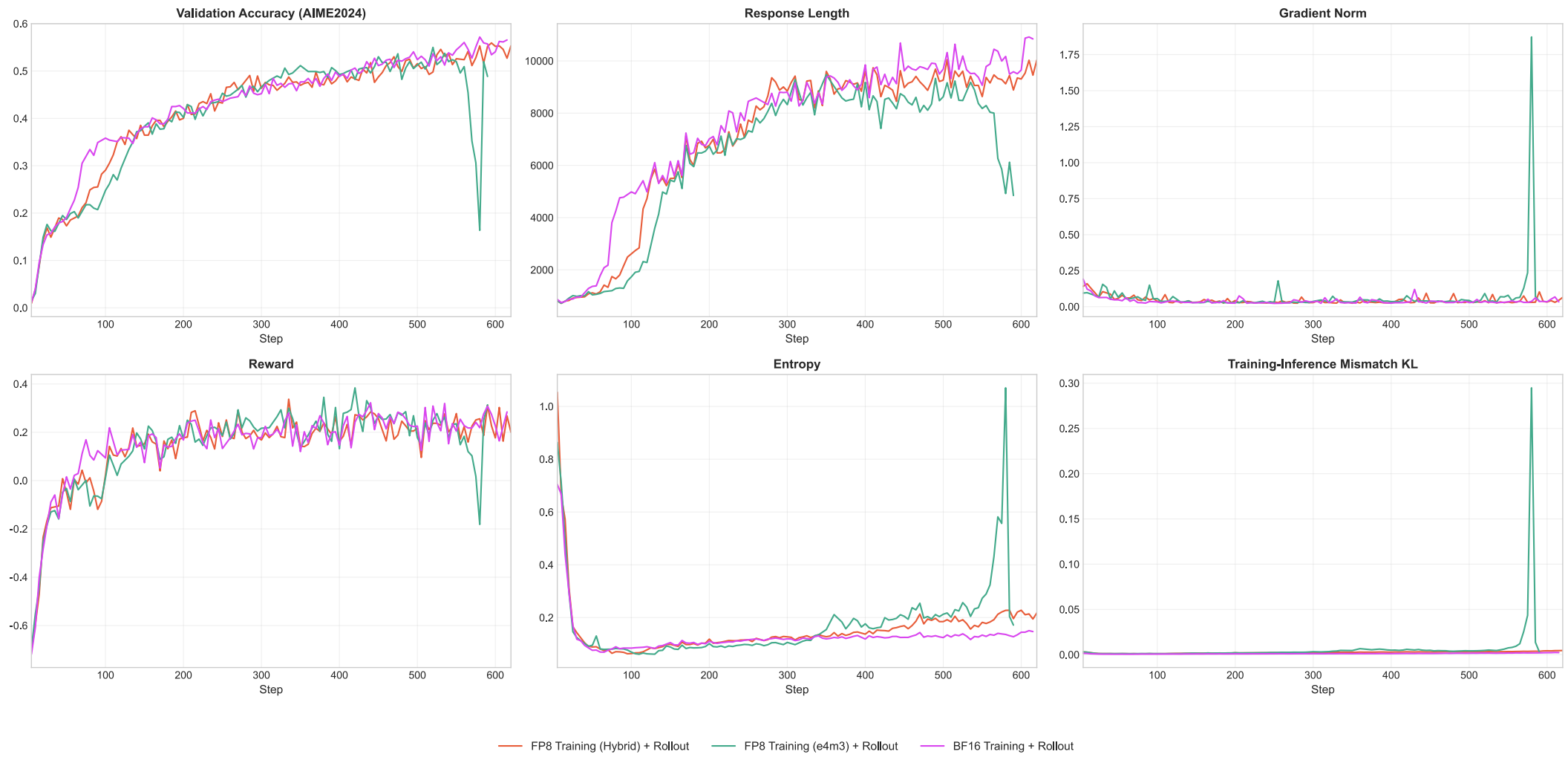

Figure 11: Ablation on FP8 training recipe for Qwen3-30B-A3B-Base. Purple: BF16 baseline. Orange: FP8 E2E with hybrid recipe (E4M3 forward + E5M2 backward). Greem: FP8 E2E with pure E4M3 recipe (E4M3 both forward and backward).

**Gradient profiling.** To quantify this, we profile the grad-output tensors across all layers under the pure E4M3 recipe. The MoE `fc1` layer is the primary problem: on average 5% of its quantization tiles exceed E4M3′s representable range, compared to less than 0.5% for `fc2`, QKV projections, and output projections. Layer 0 `fc1` is the worst at 21% exceedance. In the worst-case tiles, up to 50% of gradient data is lost to overflow or clamping. The connection to training collapse is direct: during the instability window at steps 571–576, the p99 tile exceedance rate in `fc1` doubles from 26% to 41%, confirming that gradient overflow in MoE expert layers is the proximate cause of the collapse.

While DeepSeek-V3 demonstrated that pure E4M3 can work for pretraining with careful tile-wise scaling, RL training appears more demanding due to its non-stationary optimization landscape: the policy, reward distribution, and response length all co-evolve, creating gradient dynamics that are harder to contain within E4M3′s narrower range. We therefore recommend the hybrid recipe for FP8 RL training.

**Scaling Factor: FP32 vs Power-of-2.**

The choice of scaling factor representation affects quantization accuracy and train–inference consistency. Standard FP32 scaling factors allow arbitrary values, while power-of-2 (UE8M0) scaling factors restrict scales to $2^k$ values, enabling cheaper multiply operations (bit shifts) at the cost of coarser granularity. We compare three configurations: (1) FP32 scales for both training and rollout, (2) UE8M0 scales for both training and rollout, and (3) a mixed setting with FP32 scales in training and UE8M0 scales in rollout.

Figure 12 shows the mismatch KL across the three configurations. The all-FP32 configuration (blue) consistently achieves the lowest mismatch KL, while the all-UE8M0 configuration (orange) exhibits moderately higher mismatch due to the coarser quantization granularity of power-of-2 scales.

Based on these results, we adopt FP32 scaling factors for all other FP8 end-to-end experiments in this report. While UE8M0 scales offer potential hardware efficiency advantages through bit-shift multiply, FP32 scales provide tighter train–inference alignment with negligible computational overhead on current hardware, making them the preferred default for FP8 RL training.

## 3 Related Work

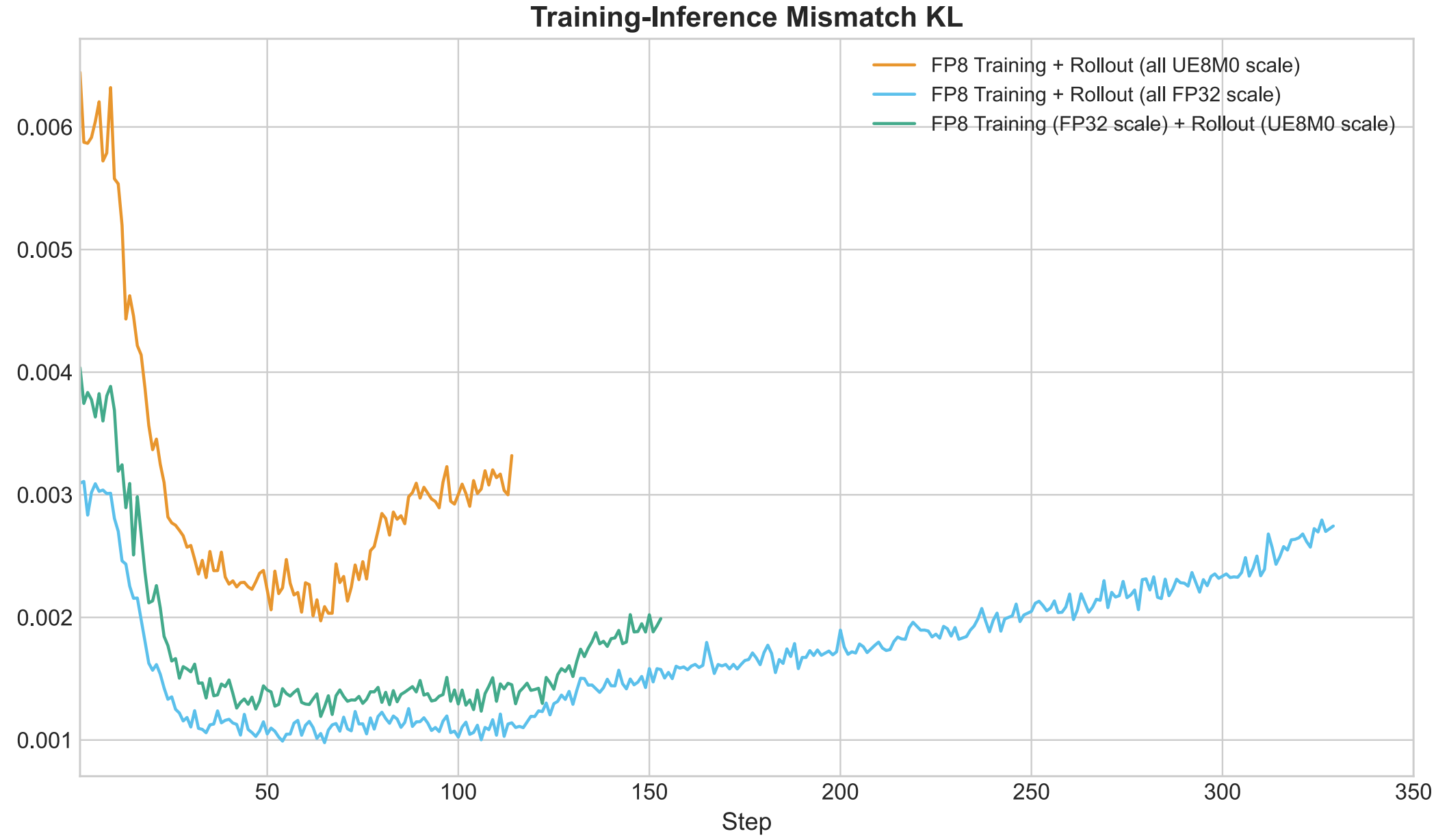

Figure 12: Ablation on scaling factor format: training-inference mismatch KL under different scaling factor configurations. Orange: FP8 E2E with all UE8M0 (power-of-2) scales. Blue: FP8 E2E with all FP32 scales. Green: FP8 training with FP32 scales + rollout with UE8M0 scales.

### 3.1 FP8 Quantization for LLMs

As models scale to trillions of parameters, low-precision formats have become essential for practical training and inference. The foundational specification for FP8 formats was established by Micikevicius et al. [12], who proposed the E4M3 and E5M2 encodings that balance dynamic range and precision. Building on this foundation, FP8-LM [13] demonstrated that FP8 can be systematically applied to weights, gradients, and optimizer states during LLM pre-training and fine-tuning, achieving 2× throughput speedup. More recently, DeepSeek-V3 [8] reported the first successful industrial-scale application of FP8 training. It employs fine-grained quantization (1×128 tiles for activations and 128×128 blocks for weights) and uses the E4M3 format for both forward and backward GEMM operations, setting a new benchmark for high-performance low-precision training.

On the inference side, practical FP8 deployment is primarily driven by high-performance serving frameworks (e.g., vLLM [3], TensorRT-LLM [2], and SGLang [4]) leveraging hardware-native support on NVIDIA Hopper and Blackwell architectures. Rather than relying on a single technique, they combine (i) reduced-precision representations to lower weight/activation and memory-traffic costs, with (ii) highly optimized attention/decoding kernels, paged KV cache management, and efficient runtime scheduling to improve throughput under long-context and high-concurrency workloads. In memory-bandwidth-bound regimes, lowering the KV cache precision can substantially reduce cache footprint, effectively increasing the number of cached tokens and the achievable concurrency, while favorable hardware–kernel support often translates these savings into meaningful end-to-end throughput gains.

### 3.2 RL Systems for LLMs

Modern RL systems for LLMs must orchestrate complex interactions between Actor, Critic, Reward, and Reference models. Early frameworks like DeepSpeed-Chat [14] utilized synchronous execution, which often led to poor resource utilization in large clusters.

Several high-performance frameworks have since emerged to address these challenges. OpenRLHF [15] utilizes a Ray-based distributed architecture to decouple model roles across GPUs and integrates vLLM for accelerated generation. verl [9] introduces a flexible programming model that allows seamless integration of various training backends (FSDP, Megatron-LM) with inference backends.

NeMo-RL [11] builds on NVIDIA's NeMo ecosystem, providing tight integration with Megatron-LM and native support for FP8 training and inference. ROLL [16] focuses on hardware-aware workload mapping for massive clusters. slime [17] provides a lightweight framework natively integrated with SGLang, specifically optimizing for routing consistency in MoE models. To overcome the inherent latency of sequential decoding, asynchronous architectures have also been proposed. AReaL [18] decouples generation and training workflows entirely, allowing rollout nodes to generate samples continuously without waiting for training updates.

### 3.3 Training-Inference Mismatch in LLM RL

Modern LLM RL stacks typically decouple rollout from training: trajectories are produced by high-throughput inference engines, while updates are computed in training backends. Even with synchronized weights, differences in numerics, kernels and parallelism can make the rollout policy deviate from the training policy, effectively turning nominally on-policy updates into off-policy ones and causing instability or collapse.

Addressing this training–inference mismatch has become a central concern. A common algorithmic approach is importance sampling (IS): it corrects off-policy updates by reweighting gradient contributions using the likelihood ratio between the training policy and the rollout policy. In practice, truncated or masked IS variants (TIS/MIS) control variance by clipping extreme weights or masking unreliable tokens/sequences [5], [7], [19]. For MoE models, Rollout Routing Replay (R3) records inference-time expert choices and replays them during training to align routing [20]. Complementary system efforts pursue bitwise consistency by aligning kernels and enforcing determinism across parallel configurations [21]. Finally, Qi et al. [22] argue a root cause can lie in floating-point precision itself: BF16 rounding error accumulates during autoregressive decoding, and using FP16 consistently across training and inference substantially reduces divergence.

## 4 Summary and Future Work

We presented FP8-RL, a practical and stable FP8 rollout stack for LLM reinforcement learning. Our system addresses the unique challenges of applying FP8 in RL—dynamic weight synchronization and train–inference mismatch—through a combination of engineering and algorithmic techniques. W8A8 linear quantization delivers 10–20% rollout speedup for dense models and 30–50% for MoE models, with KV cache quantization providing further gains by expanding effective cache capacity under memory pressure, achieving up to 44% combined improvement. End-to-end FP8, which aligns precision across both training and rollout, reduces distribution drift compared to FP8 rollout-only configurations. Across all settings, importance-sampling-based correction remains essential for maintaining accuracy comparable to BF16 baselines.

Future work includes exploring more aggressive formats such as NVFP4 (noting reported instability from accumulated quantization error), scaling validation to larger models, and extending to multi-turn and agentic RL scenarios where long contexts further stress rollout efficiency.

## A Appendix: FP8 W8A8 Linear Rollout Configuration

This appendix provides detailed configuration instructions and usage examples for enabling FP8 W8A8 linear rollout in the verl framework.

### A.1 Basic Configuration

To enable FP8 quantization, simply add the following argument to the command line:

```
actor_rollout_ref.rollout.quantization=fp8
```

With this specification, the system automatically handles the weight conversion and loading into the vLLM backend during the rollout phase.

### A.2 Complete Example

For a complete example, please refer to the DAPO Qwen3-30B FP8 Rollout Recipe. Below is a snippet showing how this is typically configured in a startup script:

```
# ... existing configuration ...
actor_rollout_ref.rollout.temperature=${temperature} \
actor_rollout_ref.rollout.top_p=${top_p} \
actor_rollout_ref.rollout.top_k=${top_k} \
+actor_rollout_ref.rollout.quantization=fp8 \
```

```
actor_rollout_ref.rollout.name=vllm \
# ... remaining configuration ...
```

### A.3 Performance Optimization Requirements

To achieve the best performance, ensure your environment meets the following requirements:

**DeepGEMM library**: The DeepGEMM library significantly accelerates FP8 GEMM operations.

- For vLLM 0.11+ and SGLang 0.55+, DeepGEMM is enabled by default.
- For vLLM 0.10.x, you must enable it manually by setting `VLLM_USE_DEEP_GEMM=True` when launching the RL training.

**CUDA version**: Must be 12.9 or higher, as newer CUDA versions provide significant speedups for FP8 DeepGEMM kernels.

### A.4 Rollout Correction Configuration

As highlighted in the main text, FP8 quantization can introduce errors that exacerbate the discrepancy between training and inference distributions. To prevent this "training-inference mismatch" from degrading training accuracy or destabilizing the RL process, it is strongly recommended to pair FP8 rollout with mismatch correction techniques, such as Token-level Importance Sampling (TIS).

The verl framework provides a comprehensive suite of functionalities to address this. For detailed guidance on configuring rollout correction, please refer to the Rollout Correction Documentation.

## B KV Cache Quantization Usage

### B.1 verl Implementation (Inference-Side Calibration)

The verl implementation leverages vLLM's native dynamic QKV scale calculation. To enable KV cache FP8:

```yaml
actor_rollout_ref:
  rollout:
    quantization:
      kv_cache_dtype: fp8_e4m3
      calculate_kv_scales: True
```

Implementation details:

- Triggers forced recalibration before each rollout phase by resetting internal `calculate_kv_scales` flags
- Requires vLLM version with FP8 KV cache support (vLLM 0.11+)
- Automatically handles per-step QKV scale updates without external intervention
- Can be combined with FP8 linear W8A8 for maximum throughput

### B.2 NeMo-RL Implementation (Trainer-Side Calibration)

The NeMo-RL implementation performs QKV scale recalibration on the training side. Configuration example:

```yaml
policy:
  generation:
    vllm_cfg:
      precision: fp8
      kv_cache_dtype: fp8
```

Implementation details:

- At the end of each training step, recalibrates QKV scales using updated policy weights

- Uses a subset of training data (prompts and generated responses) for calibration
- Synchronizes new scales to inference engine for subsequent rollout
- Calibration overhead: approximately 2-3% of total step time
- Provides fine-grained control over calibration data distribution

### B.3 NeMo-RL Experimental Validation

We validated the Trainer-Side calibration approach on Qwen3-8B-Base using the NeMo-RL framework.

#### B.3.1 Experimental Setup

The experimental setup maintains identical dataset selection and hyperparameters as specified in the main text.

**Configurations evaluated:**

- BF16 baseline (standard BF16 training)
- Linear W8A8 (W8A8 quantization for linear layers only)
- Full FP8 (FP8 quantization across all three components—linear layers (W8A8), KV cache storage, and attention computations—representing the most aggressive quantization setting.)

All FP8 configurations use token-level TIS (C=2) for rollout correction.

#### B.3.2 Training Effectiveness

Figure 13 presents the training curves across different quantization configurations. Key observations:

**Accuracy alignment**: After enabling token-level TIS, the validation accuracy of Full FP8 successfully aligns with both the BF16 baseline and Linear W8A8 configurations.

**Mismatch KL**: The mismatch KL divergence is higher when both KV cache and attention are quantized to FP8, compared to the Linear W8A8 setting. This is expected due to compounded quantization errors across multiple components.

**Consistency with Inference-Side calibration**: These observations from the Trainer-Side calibration approach are largely consistent with the Inference-Side calibration results reported in the main text, indicating that both calibration paradigms achieve comparable effectiveness in maintaining training stability under FP8 KV cache quantization.

#### B.3.3 Performance Results

Figure 14 illustrates the rollout performance across different quantization schemes. Key findings:

**Speedup magnitude**: Enabling FP8 for both KV cache and attention operations yields an additional ~30% speedup on top of the Linear W8A8 configuration, resulting in an overall ~48% speedup compared to the BF16 baseline.

**Length dependency**: Performance gains are particularly pronounced at longer response lengths, where attention computations constitute a larger fraction of the overall workload, making FP8 quantization of attention components increasingly beneficial.

**Calibration overhead**: The QKV scale recalibration process consumes approximately 2-3% of the total step time, representing a minor cost relative to the substantial rollout acceleration achieved.

These results demonstrate that Trainer-Side calibration achieves comparable training effectiveness and performance gains to Inference-Side calibration, validating both approaches as viable strategies for FP8 KV cache quantization in RL settings.

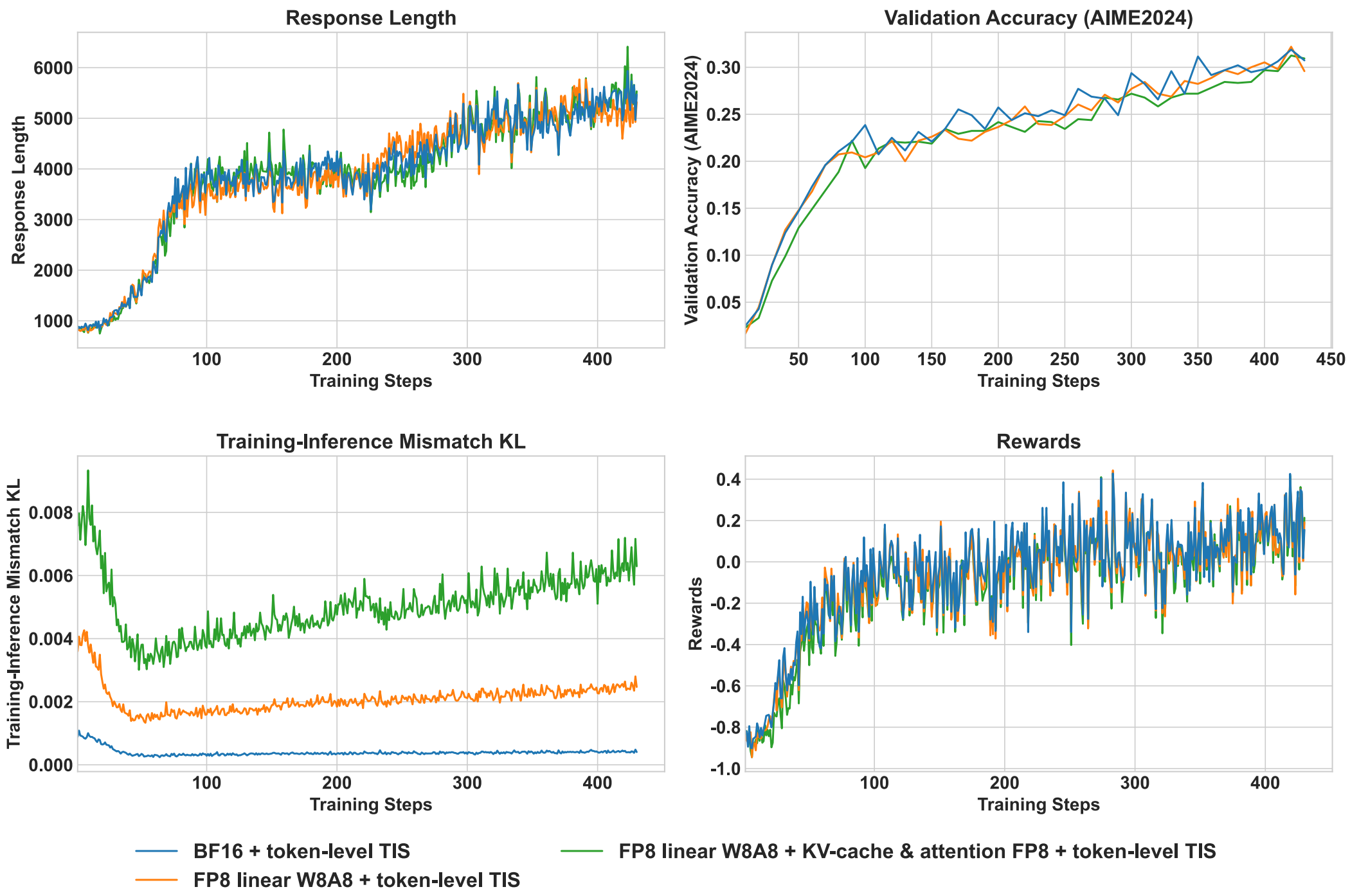

Figure 13: Training curves for Qwen3-8B-Base with Trainer-Side KV cache calibration (NeMo-RL). Blue: BF16 baseline, Orange: Linear W8A8, Green: Full FP8 (Linear + KV cache + Attention).

## C End-to-End FP8 RL: Dense Model Validation

We validate end-to-end FP8 on the Qwen3-8B-Base dense model to confirm that aligning training and rollout precision preserves learning quality and reduces mismatch.

### C.1 Experimental Setup

We evaluate end-to-end FP8 on the Qwen3-8B-Base model using the same DAPO + AIME24 validation recipe and hardware configuration as Section 2.1. All experiments are conducted on 8×H100 GPUs. The training hyperparameters follow the same settings: prompt batch size of 32 with n=16 responses per prompt (rollout batch size 32×3×16), training batch size and PPO mini-batch size both set to 32, and maximum response length of 20K tokens. Token-level TIS with C=2 is applied to all FP8 configurations for rollout correction. These experiments are implemented in the NeMo-RL framework.

**Configurations.** We compare three precision settings to isolate the impact of end-to-end FP8:

- **BF16 training + BF16 rollout**: Full BF16 baseline
- **BF16 training + FP8 rollout**: FP8 rollout-only (the same configuration as Section 2.1)
- **FP8 training + FP8 rollout**: End-to-end FP8 quantization

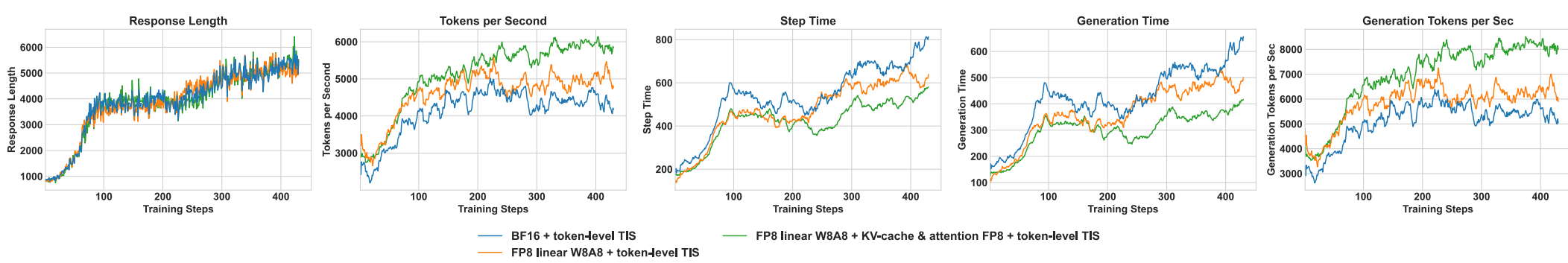

Figure 14: Rollout performance on Qwen3-8B-Base with Trainer-Side calibration (NeMo-RL). Blue: BF16 baseline, Orange: Linear W8A8, Green: Full FP8 (Linear + KV cache + Attention).

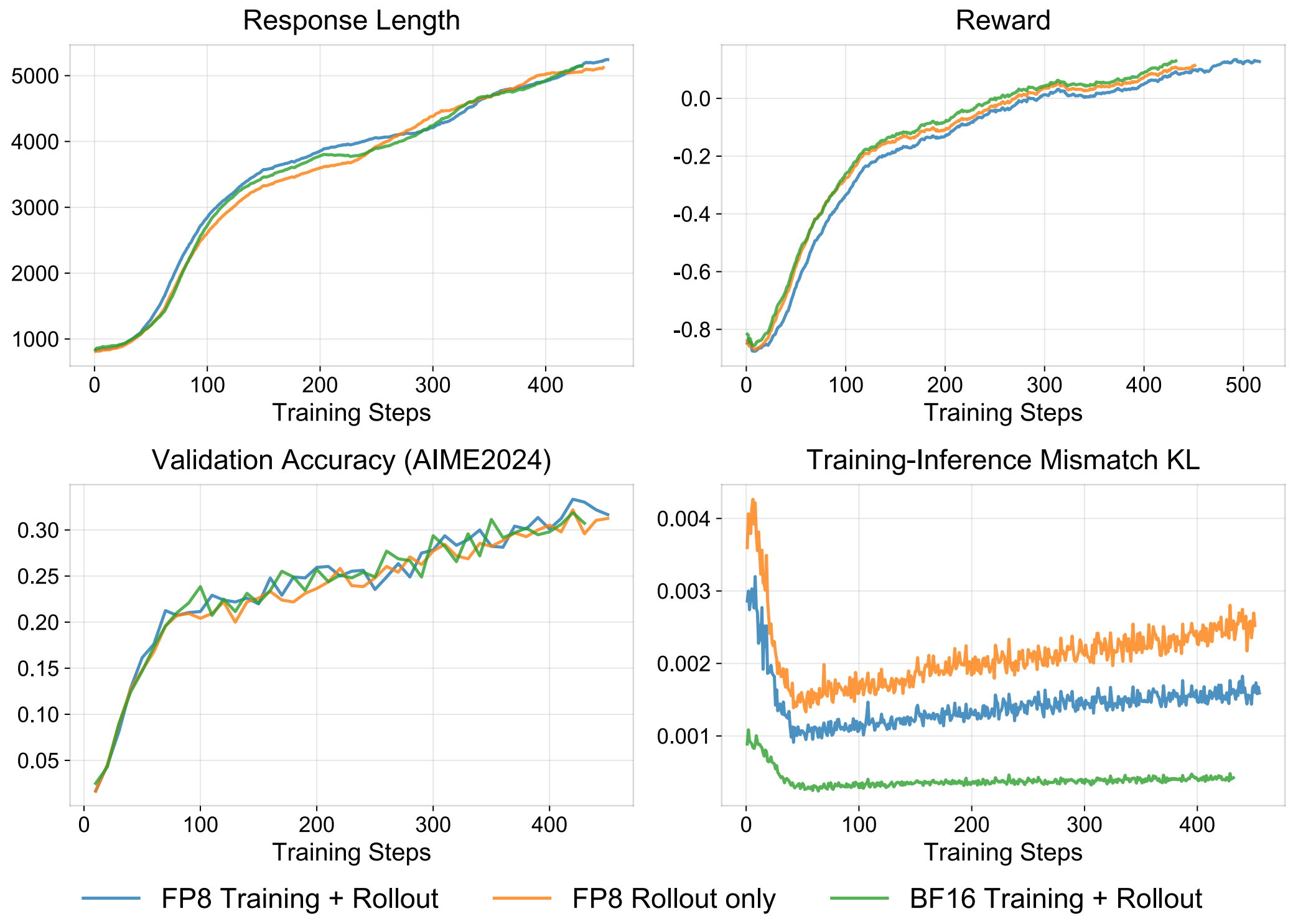

Figure 15: Training curves for end-to-end FP8 RL on Qwen3-8B-Base. Blue: FP8 training + FP8 rollout. Orange: BF16 training + FP8 rollout. Green: BF16 training + BF16 rollout.

### C.2 Results

Figure 15 presents the training dynamics across all three configurations, including validation accuracy, reward, response length, and mismatch KL.

**Accuracy alignment with BF16.** FP8 training + rollout (blue) closely tracks the BF16 training + rollout baseline (green) across response length, reward, and validation accuracy. This demonstrates that end-to-end FP8 can preserve the learning dynamics and final quality of the BF16 run under this setup.

**Reduced mismatch vs FP8 rollout-only.** Compared to FP8 rollout-only (orange), FP8 training + rollout (blue) exhibits lower training–inference mismatch KL. This confirms that aligning precision between training and rollout reduces distribution drift. However, FP8 training + rollout still shows slightly higher mismatch than BF16 training + rollout, suggesting that precision alignment helps but does not fully remove all sources of mismatch.